\def\year{2021}\relax
\documentclass[letterpaper]{article} 
\usepackage{aaai21}  
\usepackage{times}  
\usepackage{helvet} 
\usepackage{courier}  
\usepackage[hyphens]{url}  
\usepackage{graphicx} 
\usepackage{natbib}  
\usepackage{caption} 
\usepackage{tabularx}
\usepackage{booktabs}
\usepackage[mathscr]{eucal}
\usepackage{mathtools}
\usepackage{amsmath}
\usepackage{amsfonts}
\usepackage{bbold}
\usepackage{mathtools}
\DeclareMathOperator*{\argmax}{arg\,max}

\title{Nutri-bullets: Summarizing Health Studies by Composing Segments}

\author{Darsh J Shah\textsuperscript{1} ~~
Lili Yu\textsuperscript{2}~~
Tao Lei\textsuperscript{2} ~~
Regina Barzilay\textsuperscript{1}\\
\small{\normalfont{\textsuperscript{1}{Computer Science and Artificial Intelligence Lab, MIT}}}\\
\small{\normalfont{\textsuperscript{2}{\small{ASAPP, Inc.}}}\\
darsh@csail.mit.edu ~~
liliyu@asapp.com ~~
tao@asapp.com ~~
regina@csail.mit.edu}}

\begin{document}

\maketitle

\begin{abstract}
We introduce \emph{Nutri-bullets}, a multi-document summarization task for health and nutrition. 
First, we present two datasets of food and health summaries from multiple scientific studies.
Furthermore, we propose a novel \emph{extract-compose} model to solve the problem in the regime of limited parallel data. We explicitly select key spans from several abstracts using a policy network, followed by composing the selected spans to present a summary via a task specific language model. Compared to state-of-the-art methods, our approach leads to more faithful, relevant and diverse summarization -- properties imperative to this application.
For instance, on the BreastCancer dataset our approach gets a more than 50\% improvement on relevance and faithfulness.\footnote{Our code and data is available at \url{https://github.com/darsh10/Nutribullets.}}
\end{abstract}

\section{Introduction}
\begin{figure}[!t!]
 \centering
\includegraphics[width=1.0\columnwidth]{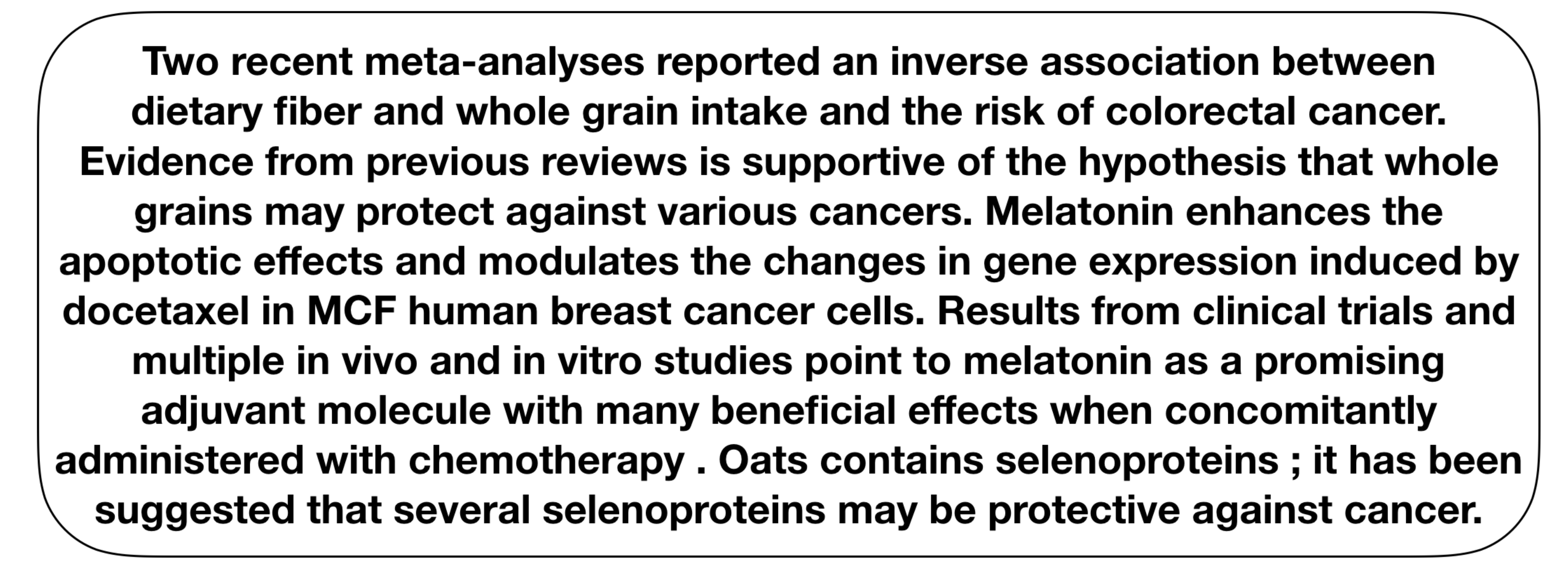}
\caption{Our model's \emph{extract-compose} summary describing impacts of whole grains on  cancer. Segments from multiple scientific studies are composed to present the output. 
}
\label{fig:example-1}
\end{figure}

Multi-document summarization is essential in domains like health and nutrition, where new studies are continuously reported (see Figure \ref{fig:example-1}). Websites like \textit{Healthline.com}\footnote{https://healthline.com}

are critical in making food and nutrition summaries available to web users and subscribers. Such summaries provide a collective view of information  -- a crucial property, especially when the consumption of such knowledge leads to health related decisions. In the current environment, it is critical to provide users with the the latest health related findings. Unfortunately, summarization is time consuming and requires domain experts.

 Through the introduction of \emph{Nutri-bullets}, a multi-document summarization task, we aim to automate this summarization  and expand the availability and coverage of health and nutrition information.

\begin{figure}[!t!]
 \centering
\includegraphics[width=1.0\columnwidth]{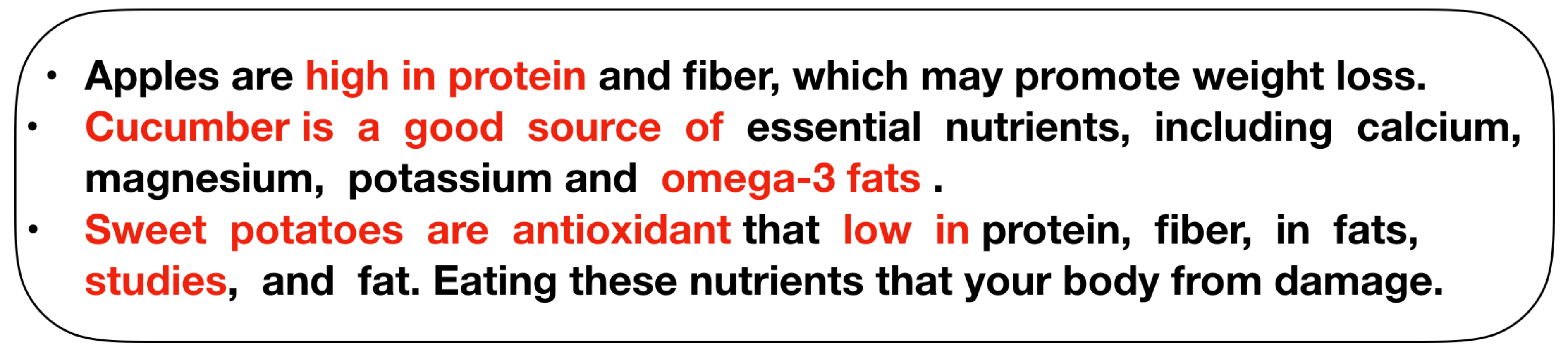}

\caption{Examples of unfaithful summaries generated by seq2seq models trained on the HealthLine dataset. Such fictitious texts (red) can be extremely misleading for readers. 
}
\label{fig:example-2}
\end{figure}

Recently, summarization has been predominantly solved by training sequence-to-sequence (seq2seq) models \cite{rush-etal-2015-neural,vaswani2017attention,hoang2019efficient,lewis2019bart}. In the multi-document setting, popular seq2seq methods either concatenate all input documents as a single source or consider a hierarchical setting \cite{liu2018generating}. While such methods generate fluent text, in our case, they fail to produce content faithful to the inputs (examples shown in Figure \ref{fig:example-2}).
Two factors make \emph{Nutri-bullets} particularly challenging for seq2seq models: (i) The concatenation of health documents constitutes long sequences with key information being scattered making composing a good summary extremely difficult; and (ii) While a vast number of scientific abstracts are available in libraries such as Pubmed\footnote{https://pubmed.ncbi.nlm.nih.gov/},
a very limited number of summaries, ranging from several hundreds to a few thousands, are available to train a summarization system.

In this paper, we present a novel and practical approach which learns to \emph{extract and compose} knowledge pieces from scientific abstracts into a summary. We side-step the scarcity of parallel data by focusing on knowledge-extraction from scientific abstracts, made possible by modules trained on such crowd-sourced annotations. Furthermore, our model learns to select a subset of knowledge pieces well-suited for a summary, from a large pool. The selection is performed through a multi-step framework, guided by reinforcement learning and distant supervision from limited available parallel data.

\begin{figure*}[!t]
\centering
\includegraphics[width=0.86\textwidth]{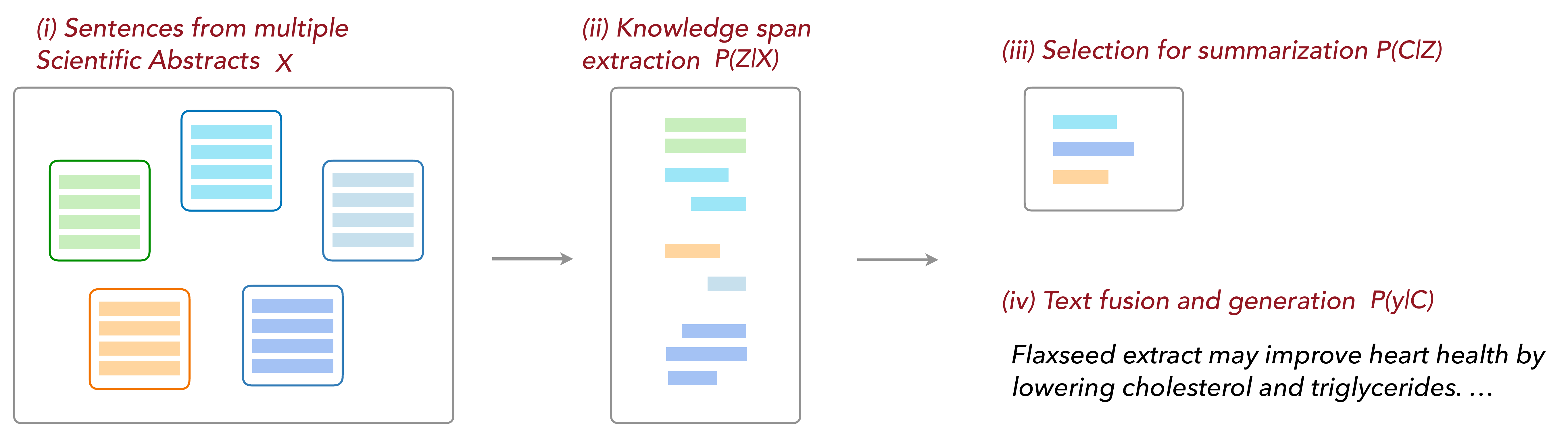}
\caption{Illustrating the flow of our extract-compose system. (i) We begin with multiple scientific abstracts to summarize from; (ii) We extract knowledge spans as possible candidates for generation; (iii) We select key spans ("improve heart health by lowering cholestrol and triglycerides" in this case) from all possible candidates and (iv) Present a summary sentence using the spans and a domain specific language model.
}
\label{fig:mask}
\end{figure*}

While considering the extracted text as inputs to a seq2seq model already makes the learning-to-generate process easier, we further explore an alternative approach to compose the extracted text into a complete summary using a text infilling language model. Once key spans are selected from different documents, they are fused using a task specific language model ~\citep{shen2020blank} as illustrated in Figure \ref{fig:mask}. The fusion of separate yet related spans through generation is critical in producing meaningful and readable summaries.\footnote{Our analysis shows that 91\% of all produced tokens are from extracted components. The generated words allow fusion and cohesion.}

We conduct human and empirical evaluation to comprehensively study the applicability and quality of these approaches. While seq2seq models enjoy high fluency scores, the \emph{extract-compose} method performs much stronger on metrics such as content, relevance, faithfulness and informativeness.
Our method is particularly dominant in scenarios with scarce parallel data, since our model requires little summary data for training. 
For instance, on the BreastCancer dataset, humans rate the \emph{extract-compose} model's summaries \textbf{more than 50\% higher} for relevance and faithfulness than the next best baseline. Comparison with strong baselines and model ablation variants highlights the necessity of a distant supervision and reinforcement learning based multi-step approach, in selecting key ordered spans amongst several possible combinations, for text composition. Our contributions 
are threefold:
\newline
(i) We collect two new nutrition and health related datasets for multi-document summarization. We also collect large-scale knowledge extraction annotations, applicable to numerous tasks. \newline
(ii) We demonstrate the effectiveness of our modelling approach for generating health summaries given limited parallel data. Our approach strongly outperforms all baselines and variants on human and automatic evaluation. \newline
(iii) We conduct comprehensive (human and automatic) evaluation focusing on content relevance, faithfulness and informativeness -- metrics more relevant to the task. These set new benchmarks to critically evaluate summaries in high impact domains.

\section{Related Work}
\textbf{Multi-document Summarization}
Approaches in neural sequence-to-sequence learning~\citep{rush-etal-2015-neural, cheng-lapata-2016-neural,see-etal-2017-get} for document summarization have shown promise and have been adapted successfully for multi-document summarization~\citep{zhang-etal-2018-adapting, lebanoff2018adapting, baumel2018query, amplayo2019informative, multinews}. Trained on large amounts of data, these methods have improved upon traditional extractive~\citep{carbonell1998use, radev-mckeown-1998-generating, haghighi2009exploring} and abstractive approaches~\citep{barzilay1999information, mckeown1995generating,ganesan2010opinosis}. Despite producing fluent text, these techniques also tend to generate false information which is not faithful to the original inputs~\citep{puduppully2019data,kryscinski2019evaluating}. Side-information, such as citations in scientific domains~\citep{qazvinian2008scientific,qazvinian2013generating} or semantic representations~\citep{liu-etal-2015-toward}, can be used to improve this \cite{sharma-etal-2019-entity, wenbo2019concept, puduppully2019data, koncel2019text}. However, such methods struggle in low resource scenarios. In this work, we are interested in producing faithful and fluent text in a technical domain where few parallel examples are available.

\textbf{Text Fusion}
Traditionally, sentence fusion approaches~\citep{barzilay2005sentence} aid the concatenation of different text fragments for summarization. Recent language modeling approaches like \citet{devlin2018bert, stern2019insertion} can also be extended for completion and fusion of partial text. These models have more flexibility than those trained on text fusion datasets~\citep{narayan-etal-2017-split, geva2019discofuse} that can combine two fragments only. In this work, we modify the Blank Language Model \citep{shen2020blank} to combine fragments coming from different source documents. 

\textbf{Deep Reinforcement Learning for Text Summarization}
The inherent discrete and sequential nature of text generation tasks has made optimizing generation models with reinforcement learning~\citep{sutton2018reinforcement} very popular~\citep{keneshloo2019deep}. Typically, automatic evaluation metrics, like BLEU~\citep{papineni2002} or ROUGE~\citep{lin2004}, are used to provide the reward signal to train reinforcement learning algorithms in translation~\citep{wu2018study}, summarization~\citep{paulus2017deep, pasunuru-bansal-2018-multi,chen-bansal-2018-fast,xu2019clickbait} and question generation~\citep{zhang2019addressing} tasks. In this work, we are interested in using reinforcement learning to iteratively, select an explicit \emph{ordered} subset of text phrases for downstream fusion. 

\section{Method}
 In this section, we describe the framework of our \emph{extract-compose} solution for \emph{nutri-bullets}. Our goal is to produce a text summary $y$ for a food item from a pool of multiple scientific abstracts $X$.

\subsection{Extract-compose Framework}
We attain food health entity-entity relations, for both input documents $X$ and the summary $y$, from entity extraction and relation classification modules trained on corresponding annotations(Table \ref{table:entity_relations}).

For input documents, we collect $\{(x_p, \mathcal{G}_p)_{p=1}^{N}\}$, where $x_p$ is the sentence text and  $\mathcal{G}_p$ is the set of entity-entity relations, in $N$ sentences and $X=\{x_p\}$ is the raw text. $\mathcal{{G}}_p = \{(e_i^k,e_j^k,r^k)^K\}$ is composed of $K \in \{0, 1, 2, \dots\}$ tuples of two entities $e_i$, $e_j$ and their relation $r$. $r$ represents relations such as the effect of a nutrition entity $e_i$ on a condition $e_j$ (see Table \ref{table:entity_relations}). \footnote{We train an entity tagger and relation classifier to predict $\mathcal{G}$ and also for computing knowledge based evaluation scores.}

Similarly, we denote the summary data as $\{(y_q, \mathcal{G}_q)_{q=1}^{M}\}$, where $y_q$ is a concise summary and $\mathcal{G}_q$ is the set of entity-entity relation tuples, in $M$ summaries.

Joint learning of content selection, information aggregation and text generation for multi-document summarization can be challenging. This is exacerbated in our technical domain with few parallel examples.
We propose to overcome these challenges by first learning to extract and then composing the informative text pieces. 

We model two intermediate variables, distilled text spans $Z$ associated with all entity-entity relation information $\mathcal{G}$, and the key content $C$ selected from them.

The probability of an output summary $y$, conditioned on the input $X$ is 
\begin{equation}
        P(y|X) = \sum_{C, Z}P(y|C)P(C|Z)P(Z|X)
\end{equation}

Where: (i) $P(Z|X)$ models a knowledge extraction process  gathering all entity-entity relations and corresponding text spans, $\mathcal{G}$; (ii) $P(C|Z)$ models the process of selecting an important subset from all extracted text spans and (iii) $P(y|C)$ models the text fusion process that composes fluent text incorporating the extracted content.

\subsection{Span Extraction $P(Z|X)$}
We model $P(Z|X)$ as a span extraction task leveraging a rationale extraction system~\citep{lei-etal-2016-rationalizing}.
The extraction model picks the fewest words from the input necessary to make the correct relation label prediction for an entity pair.
Let $(e_i,e_j, r)$ be one entity relationship tuple, $x$ be the associated sentence text and $z$ be the considered rationale text span.
$P(Z|X)$ is trained to minimize the loss: 
\begin{align}
    \mathcal{L} = \mathcal{L}(z,r)+ \mathcal{L}(z) + \mathcal{L}(z, e_i, e_j)
\end{align}.
Where $\mathcal{L}(z,r)$ is the cross entropy loss for predicting $r$ with $z$ as the extracted text. 
$\mathcal{L}(z)$ is a regularization term to select short and coherent text, by minimizing the span lengths and discontinuities among the spans. In addition to prior work, we introduced $\mathcal{L}(z, e_i, e_j)$ to encourage the selection of phrases that contain entities $e_i$ and $e_j$. Specifically, we construct verb phrases (from constituency parse trees) containing the condition entity, and minimize the distance between the selected text span $z$ and the verb phrase.  Empirically, this loss stabilizes the span extraction and improves the quality of selected text spans, using $r$ labels as indirect supervision as in ~\citet{shah2019automatic}.
By running the extraction model on every $(e_i,e_j,r)$ tuple, we distill the input text into a set of text spans $Z =\{z_1, z_2, \dots, z_m\}$. 
\subsection{Content Selection Policy Network $P(C|Z)$}
$P(C|Z)$ is a policy network,
that takes a large set of text spans $Z$ as input, and outputs $C$, an \emph{ordered} set of key text spans. 
We model our content selection as a finite Markov decision process (MDP). Where the state is represented as $s_t= (t,\{c_1, \dots, c_t\}, \{z_1, z_2, ..., z_{m-t}\})$ for step $t$, content selected so far $\{c_1, \dots, c_t\}$ and remaining text spans $\{z_1, z_2, ..., z_{m-t}\}$.
Our action space is all the remaining text spans plus one special token, $Z\cup \{\textit{STOP}\}$. The number of actions is $|m-t|+1$. 
We parameterize the policy $\pi_\theta (a | s_t)$ with a neural network to map the state $s$ to a probability distribution over all available actions. At each step, the probability that the policy selects $z_i$ as a candidate is:
\begin{equation}
\pi_\theta (a\!=\!z_i|s_t) = \frac{\exp(f(t,\hat{z_i}, \hat{c_i*}))}{\sum_{j=1}^{m-t+1}\exp(f(t, \hat{z_j}, \hat{c_j*}))}
\end{equation}
where $c_i* = \argmax_{c_j}(cos(\hat{z_i},\hat{c_j}))$ 
is the selected content closest to $z_i$, $\hat{z_i}$ and 
$\hat{c_i*}$ 
are the encoded dense vectors, 
$cos(u,v)=\frac{u \cdot v}{||u||\cdot ||v||}$ 
and $f$ is a feed-forward neural network that outputs a scalar score. 
The selection process begins from $Z$. Our module iteratively samples actions from $\pi_\theta(a|s_t)$. On picking \textit{STOP}, we end with the selected content $C$ and a corresponding reward. 
Our rewards guide the selection of informative, diverse and readable phrases:
\begin{itemize}
\setlength\itemsep{0.005em}
    \item $\mathcal{R}_e = \sum_{c \in C}cos(\hat{e_{ic}}, \hat{e_{iy}}) + cos(\hat{e_{jc}}, \hat{e_{jy}})$ is the cosine similarity of the structures of the selected content $C$ with the structures present in the gold summary $y$ (each gold summary structure accounted with only one $c$), encouraging the model to select relevant content.
    \item $\mathcal{R}_d = \mathbb{1}[\max_{i,j}(cos(\hat{c_j},\hat{c_i})) < \delta] $ computes the similarity between pairs within  selected content $C$, encouraging the selection of diverse spans.
    \item $\mathcal{R}_s\!=\! \sum_{c}\mathbb{1}[\sum\limits_{q=1}^{M}cos(\hat{c},\hat{y_q}) > \sum\limits_{q=1}^{M}cos(\hat{z'},\hat{y_q})]$, where $z'$ is a text span randomly sampled from scientific abstract corpus, predicts if $c$ is closer to a human summary than a random abstract sentence, encouraging the selection of concise spans and ignoring numerical details. 
    \item $\mathcal{R}_m$ is the Meteor \citep{denkowski2014meteor} score between the final selection with the gold summary, a signal which can only be provided at the completion of the episode.
    \item $r_p$ is a small penalty for each action step. 
\end{itemize}
The final multi-objective reward is computed as
\begin{equation}
\mathcal{R} = w_e\mathcal{R}_e \! +\!  w_d\mathcal{R}_d + w_s \mathcal{R}_s + w_m \mathcal{R}_m 
    - |C|r_p , 
\end{equation}
where, $w_e$, $w_d$, $w_s$, $w_m$ and $r_p$ are hyper-parameters. 
During training, the network is updated with these rewards. Our paradigm allows an exploration of different span combinations while incorporating delayed feedback. 

\begin{table*}[t!]
\centering
\footnotesize
\begin{tabular}{l|c|c|c|c}
\toprule
 Relation Type & \bf $e_i$      & \bf $e_j$    & \bf $r$& \bf Example   \\ 
\midrule
Containing & Food, Nutrition  & Nutrition   & Contain          & (apple, fiber, contain)               \\
\midrule
Causing  & \begin{tabular}[c]{@{}c@{}}Food, Nutrition, \\ Condition\end{tabular} & Condition   & \begin{tabular}[c]{@{}c@{}}Increase, Decrease, \\ Satisfy, Control\end{tabular}    & \begin{tabular}[c]{@{}c@{}}(bananas, metabolism, increase), \\ (orange juice, hydration, satisfy)\end{tabular} 
\\ 
\bottomrule
\end{tabular}
\caption{Details of entity-entity relationships that we study and some tuple examples.}
\label{table:entity_relations}
\end{table*}

\subsection{Text Fusion \textbf{$P(y|C)$}}
The text fusion module, $P(y|C)$, composes a complete summary using the text spans $C$ selected by the policy network. 

We propose to utilize the recently developed Blank Language Model (BLM)~\citep{shen2020blank}, which fills in the \textit{blanks} by iteratively determining the word to place in a \textit{blank} or adding a new \textit{blank}, until all \textit{blanks} are filled. The model is trained on the WikiText-103 dataset \citep{merity2016pointer}.

We extend this model with additional categorical \textit{blanks} between different text spans in $C$, according to their relation type. This ensures control over facts and relations captured from the scientific abstracts $X$ to present a semantically valid and fluent summary (details see Appendix). 

 In the Transformer variant of the model, we train a seq2seq $P(y|C)$ using the limited parallel data.

\section{Data}
In this section, we describe the dataset collected for our \emph{Nutri-bullet} system.

\begin{table}[!t]
\centering
\small
\scalebox{0.9}{
\begin{tabular}{l|c|c|c}
\hline
\bf Data    & \bf Train & \bf Dev & \bf Test      \\ \hline
Input Scientific Abstracts & 6110 & 750 & 866 \\
Average words & 327.7 & 323.3 & 332.3 \\ 
\hline

HealthLine summaries  & 1522 & 179 & 193 \\
Average words & 24.7 & 23.3 & 23.9 \\ 
Abstracts Per Instance & 3.19 & 2.82 & 3.43\\
\hline
BreastCancer summaries & 104 &18  &19 \\

Average words & 161.4 & 131.6 & 148.1  \\
Abstracts Per Instance & 18.90 & 17.21 & 17.67\\
\hline
\end{tabular}
}
\caption{Statistics for scientific abstracts, HealthLine and BreastCancer datasets.}
\label{table:splits}
\end{table}

\subsection{Corpus Collection} Our Healthline\footnote{https://www.healthline.com/nutrition} and BreastCancer\footnote{https://foodforbreastcancer.com/} datasets consist of scientific abstracts as inputs and human written summaries as outputs.

\textbf{Scientific Abstracts}  We collect 7750 scientific abstracts from Pubmed and the ScienceDirect, each averaging 327 words. The studies in these abstracts are cited by domain experts when writing summaries in the Healthline and BreastCancer datasets.  A particular food and its associated abstracts are fed as inputs to our \emph{Nutri-bullet} systems. We exploit the large scientific abstract corpus when gathering entity, relation and sentiment annotations (see Table \ref{table: annotation-stats}) to overcome the challenge of limited parallel examples. Modules trained on these annotations can be applied to any food health scientific abstract.

\textbf{Summaries} Domain experts curate summaries for a general audience in the Healthline and BreastCancer datasets. These summaries describe nutrition and health benefits of a specific food (examples shown in Appendix
). In the HealthLine dataset, each food has multiple bullet summaries, where each bullet typically talks about a different health impact (hydration, anti-diabetic etc). In BreastCancer, each food has a single summary, describing in great detail its impact on breast and other cancers. 

\textbf{Parallel Instances} The references in the human written summaries form natural pairings with the scientific abstracts. We harness this to collect 1894 parallel (abstracts, summary) instances in HealthLine, and 141 parallel instances in BreastCancer (see Table \ref{table:splits}).  Summaries in HealthLine average 24.46 words, created using an average of 3 articles.  Summaries in BreastCancer have an average length of 155.71 words referencing an average of 18 articles. Unsurprisingly, BreastCancer is the more challenging of the two.

\subsection{Entity, Relation and Sentiment Annotations}
Despite having a small parallel data compared to \citet{hermann2015teaching,narayan2018don},
we conduct large-scale crowd-sourcing tasks to collect entity, relation and sentiment annotations on Amazon Mechanical Turk. 
   The annotations (see Table \ref{table: annotation-stats}) are designed to capture the rich technical information ingrained in such domains, alleviating the difficulty of multi-document summarization and are broadly applicable to different systems \citep{koncel-kedziorski-etal-2019-text}.

\textbf{Entity and Relation Annotations} Workers identify \textit{food}, \textit{nutrition}, \textit{condition} and \textit{population} entities by highlighting the corresponding text spans. 

Given the annotated entities in text, workers are asked to enumerate all the valid relation tuples $(e_i, e_j, r)$. Table \ref{table:entity_relations} lists possible combinations of $e_i$, $e_j$ and $r$ for each relation type, along with some examples.

The technical information present in our domain can make annotating challenging. To collect reliable annotations, we set up several rounds of qualification tasks \footnote{To set up the qualification, the authors first annotate tens of examples which serve as gold answers. We leverage Mturk APIs to grade the annotation by comparing with the gold answers.}, offer direct communication channels to answer annotators' questions and take majority vote among 3 annotators for each data point. 
As shown in Table~\ref{table: annotation-stats}, we collected 91K entities,  34K pairs of relations, and 7K sentiments\footnote{For BreastCancer, we use the entity tagger and relation classifier finetuned on scientific abstracts and HealthLine datasets to extract the entities and relations.}.
The high value of mean Cohen's $\kappa$ highlights high annotator agreement for all the tasks, despite various challenges.

\begin{table}[!t]
\centering
\scalebox{0.8}{
\begin{tabular}{l|c|c|c}
\toprule
\bf Data    & \bf Task                & \bf \# annotations & \bf mean $\kappa$       \\ 
\midrule
Scientific  & entity    &  83543          & 0.75           \\
Abstracts  & relation   &  28088            & 0.79, 0.81 \\ 
  & sentiment           & 5000          &          0.65      \\ 
\midrule
HealthLine & entity     & 7860           & 0.86           \\ 
  & relation  & 5974            & 0.73, 0.90 \\ 
 & sentiment           &      2000       &         0.89       \\ 
\bottomrule
\end{tabular}
}
\caption{Entity, relation and sentiment annotation statistics. Each annotation is from three annotators. Mean $\kappa$ is the mean pairwise Cohen's $\kappa$ score.}
\label{table: annotation-stats}
\end{table}
\textbf{Food Health Sentiment}
Additionally, we also collect food sentiment annotations for evaluating our system. Food health sentiment (\textit{positive}, \textit{negative}, \textit{neutral}) indicates the kind of impact the food has on human health. The annotation is performed at a sentence level, and modules trained on this data are used to assess the contrastiveness in a food's summary bullets.

Annotation interfaces, instructions as well as more data details can be found in Appendix.
\section{Experimental Setup}
\textbf{Datasets} We randomly split both HealthLine and BreastCancer datasets into training, development and testing sets(see Table~\ref{table:splits}). 
\begin{figure*}[t!]
\centering
\includegraphics[width=0.87\linewidth]{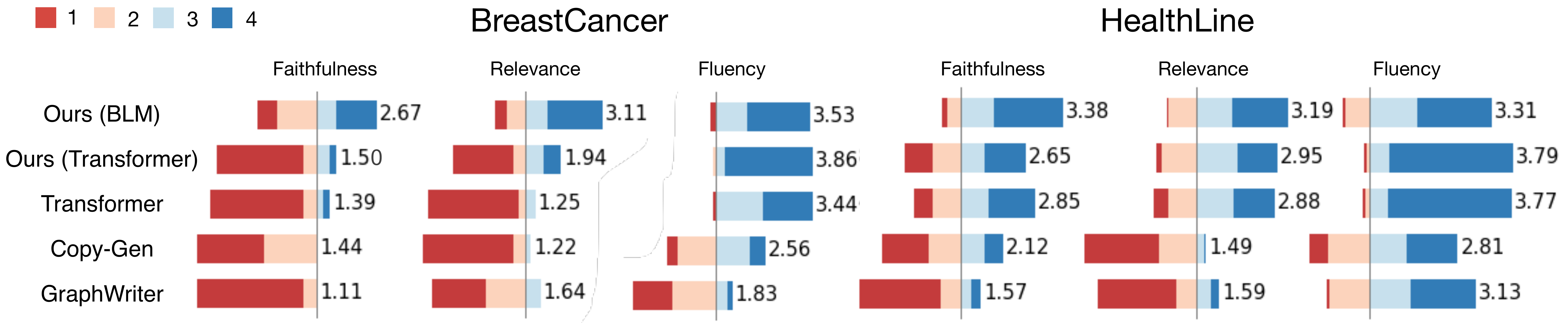}
\caption{Human evaluation Gantt charts showing ratings on faithfulness, relevance and fluency for both BreastCancer and HealthLine datasets, where Ours (BLM) -- \emph{extract-compose} model strongly outperforms all others. Each example is rated by three judges and the mean is reported.}
\label{fig:humaneval}
\end{figure*}

\textbf{Evaluation} 
The subjective nature of summarization demands human judgements to comprehensively evaluate model performance. We have human annotators score our models on faithfulness, relevance, fluency and informativeness. Given input scientific abstracts, \textit{faithfulness} characterizes if the output text is consistent with the input. Given a reference summary, \textit{relevance} indicates if the output text shares similar information. \textit{Fluency} represents if the output text is grammatically correct and written in well-formed English. \textit{Informativeness} characterizes the degree of health related knowledge conveyed by the summaries.\footnote{On the HealthLine datasets, each food article contains multiple bullet summaries. We group these bullet summaries per food for annotator comparison. For BreastCancer, a single summary output is a food's complete summary.} 

Annotators rate faithfulness, relevance and fluency on a 1-4 scale likert score, which is commonly used as a standard psychometric scale to measure responses in social science and attitude research projects~\cite{croasmun2011using,li2013novel,sullivan2013analyzing}. For rating informativeness, we perform a one-vs-one evaluation between our model and the strong Transformer baseline~\citep{hoang2019efficient}. We have 3 annotators score every data point and take an average across the scores.

We further evaluate our systems using the following automatic metrics. \textit{Meteor} is an automatic metric used to compare the model output with the gold reference. \textit{KG(G)} computes the number of entity-entity pairs with a relation in the gold reference, that are present in the output.\footnote{We run entity tagging plus relation classification on top of the model output and gold summaries. We match the gold $(e_i^{g},e_j^{g},r^{g})$ tuples using word embedding based cosine similarity with the corresponding entities in the output structures $(e_i^{o},e_j^{o},r^{o})$. If the cosine score exceeds a threshold of 0.7, a match is found.} This captures relevance in context of the reference.
\textit{KG(I)}, similarly,  computes the number of entity-entity pairs in the output that are present in the input scientific abstracts. This measures faithfulness with respect to the input documents.
\textit{Diversity} calculates the proportion of unique trigrams in the outputs \citep{li2016deep,rao2019answer}. 
\textit{Contrastiveness} calculates the proportion of article summaries belonging to the same food that contain both positive and negative/neutral sentiment about a food's health impact. 

\textbf{Baselines}
We compare our method against several state-of-the-art methods.
\begin{itemize}
\setlength\itemsep{0.02em}
    \item Copy-gen: \citet{see-etal-2017-get} is a top performing technique for summarization, which can copy words from the input or generate new words.
    \item Transformer: \citet{hoang2019efficient} is a system that utilizes a pretrained Transformer for summarization.
    \item GraphWriter: \citet{koncel-kedziorski-etal-2019-text} is a graph transformer based model, which generates text using a seed title and a knowledge graph. This model utilizes all the extraction capabilities used by our $P(Z|X)$ module.
    \end{itemize}
 In addition, we study variants of our model (1) replacing BLM with a transformer text fusing model -- Ours (Transformer),  (2) replacing the policy gradient selector with a fully supervised selector trained on knowledge structure information (Select w/ Sup.), and (3) different reward combinations.
 Further, we compare our system with a BERT-based Extractive Summarization system.

 \textbf{Implementation Details}
We adapt the implementation of span extraction from \citet{bao2018deriving}. Our policy network is a three layer feedforward neural network. We set $w_e$, $w_d$ and $w_s$ to 1 and $w_m$ to 0.75. $r_p$ is 0.02 and $\delta$ is 0.99. We use a large, 6 layer BLM \citep{shen2020blank} for fusion. Additional experimentation details can be found in the Appendix . 

We train a Neural CRF tagger \citep{yang2018ncrf++} for the food, condition and nutrition entity tagging. We use BERT \citep{devlin2018bert} text classifiers to predict the relation between two entities, and to predict food sentiments of a summary,  trained using annotations from Table \ref{table: annotation-stats}. 

\section{Results}
In this section, we describe the performance of \emph{Nutri-bullet} systems (baselines and models) on the HealthLine and BreastCancer datasets. We also report ablation studies, highlighting the importance of our policy network's ability to explore different span combinations, while incorporating long term rewards from distant supervision.
\begin{table}[]
\small
\centering
\begin{tabular}{l|c|c}
\toprule
        & BreastCancer & HealthLine \\
\midrule
Better &  94 \%        &  75  \%      \\
Worse  &  6 \%          & 25   \%    \\
\bottomrule
\end{tabular}
\caption{Human evaluation on informativeness of Ours (BLM) when comparing to Transformer.}
\vspace{-5pt}
\label{tab:human-eval-AvsB}
\end{table}

\begin{table*}[!htbp]
\centering
\footnotesize
\label{tab:casestudy}
\begin{tabularx}{\textwidth}{p{1.8cm}X}
    \toprule
    Food     & banana \\
     Pubmed Titles  & (1) High content of dopamine  a strong antioxidant, in Cavendish banana; (2) Flavonoid intake and risk of CVD: a systematic review and meta-analysis of prospective cohort studies; (3) Antioxidants in health, disease and aging; (4) Banana(Musa spp) from peel to pulp: ethnopharmacology, source of bioactive compounds and its relevance for human health. \\
    \midrule
\end{tabularx}
\begin{tabularx}{\textwidth}{p{1.8cm}Xl}
    \textit{Gold} & \textit{Bananas are high in several antioxidants, which may help reduce damage from free radicals and lower your risk of some diseases.}  \\
    Copy-gen & Seaweed contain several compounds that can reduce feelings of fullness and prevent fullness hormones, that are needed to health health benefits. \\
    GraphWriter & The type of insoluble fiber found in banana banana may help reduce blood sugar levels and help lower blood pressure.\\
    Transformer & The antioxidants in walnuts may help lower blood pressure in people with type 2 diabetes.  \\
    Ours (Trans.)  & Bananas are a rich source of antioxidants, which may help reduce the risk of many chronic diseases.  \\
    Ours (BLM)          & Bananas are used to help overcome or prevent a substantial number of illnesses , as depression and bananas containing antioxidants may lower the incidence of disease , such as certain cancers , cardiovascular and neurodegenerative diseases , DNA damage , or even have anti-aging properties .\\
    \bottomrule
\end{tabularx}
\vspace{-5pt}
\caption{Example outputs for a HealthLine input.}
\end{table*}

\begin{table*}[!htbp]
\small
\centering
\scalebox{1.0}{
\begin{tabular} {l|c|c|c|c|c|c|c|c|c|c} 
\toprule
   & \multicolumn{5}{c|}{ {BreastCancer}} & \multicolumn{5}{c}{ {HealthLine} }    \\
\midrule
Model    & Me     & KG(G) & KG(I) & Di    & Co    &  Me   & KG(G) & KG(I)   & Di     & Co   \\
\midrule
Copy-gen             & 14.0  & 21   & 23   & 70  & 0\textsuperscript{\textdagger}    & 7.4   & 21 & 51 & 82 & 43 \\
GraphWriter         & 10.1  & 0 \textsuperscript{\textdagger}& 0\textsuperscript{\textdagger} & 16  & 0\textsuperscript{\textdagger}  & 7.6   & 3 & 69 & 31  & 25 \\
Transformer          & 13.0  & 31  & 11   & 76 & 66  & 10.2  & 21 & 67 & 53 & 28 \\
\midrule
Ours (Tranformer) & \textbf{15.0}    & 49 & 13 & 81  & 50  & \textbf{10.3}  & 23 & 64 & 55 & 28 \\
Ours (BLM)        & 13.8  & \textbf{52}  & \textbf{96}  & \textbf{91}  & \textbf{94} & 8.7    &\textbf{40} & \textbf{88} & \textbf{90} & \textbf{83} \\
\bottomrule
\end{tabular}
}
\caption{Meteor score (Me), KG in gold(G), KG in input(I), Diversity (Di) and Contrastiveness (Co) in our models and various baselines, on both BreastCancer and HealthLine datasets. \textsuperscript{\textdagger}denotes cases which model generates meaningless results due to small training size. The best results are in bold and Ours(BLM) -- \emph{extract-compose} is the most dominant.}
\label{tab:automatic-full}
\end{table*}

\textbf{Human Evaluation}
Figure ~\ref{fig:humaneval} and Table~\ref{tab:human-eval-AvsB} report the human evaluation results of our model and baselines on the test set of both datasets. 

Our method outperforms all other models by a large margin on faithfulness, owing to the extraction and controlled text generation process. Our method is also rated the highest on relevance (3.11 \& 3.19), demonstrating its ability to select the important and relevant content consistently, even with little supervision. Our model produces fluent text. 

On the contrary, despite being fluent, transformer models fail to generate faithful content. They achieve decent averaged relevance scores ($\approx$2.9) on the HealthLine dataset. However, annotators rate them a \textit{score 1} far more often,\footnote{Annotators rate transformer with \textit{score 1} 6 times and ours(transformer) 3 times more often than BLM model.} due to hallucinated content and factually incorrect outputs. 

Copy-gen and GraphWriter struggles to score high on either of our metrics, showing the challenges of learning meaningful summaries in such a low-resource and knowledge-rich domain.

Finally, we evaluate if our output text can benefit the reader and provide useful nutritional information about the food being described. In our human evaluation of informativeness, we outperform 94\%-6\% and 75\%-25\% against the Transformer baseline on BreastCancer and HealthLine respectively. 

\textbf{Automatic Evaluation}
Table ~\ref{tab:automatic-full} presents the automatic evaluation results on BreastCancer and HealthLine datasets.

High KG(I) (52\% \& 40\%) and KG(G) (96\% \& 88\%) scores for our method highlight that our produced text is faithful and relevant, consistent with human evaluation. Additionally, high diversity (91\% \& 90\%) and contrastiveness (94\% \& 83\%) scores indicate that our model is also able to present distinct (across sentiment and content) information for the same food. 

In contrast, the sequence-to-sequence based methods tend to get a higher Meteor score with a lower diversity, suggesting that they repeatedly generate similar outputs, regardless of the input text. Low KG scores show that they fail to capture the relevant or faithful text, which is crucial in this domain.  Among these, our transformer variation performs strong, especially on the BreastCancer dataset. This was also observed in \citet{liu2019text}, where abstractive summarization fed with only a subset of the input sentences outperformed vanilla abstractive summarization.

\textbf{Case Study}
Table 5 shows examples of various system outputs on HealthLine. The food name and titles of corresponding scientific abstracts are also presented. We observe that Transformer, GraphWriter and Copy-gen fail to generate meaningful summaries that are faithful to the original input text. With carefully selected knowledge, our (transformer) and our (BLM) both produce relevant and useful information. Ours (transformer) generates concise and easier language.  Ours (BLM) composes faithful text with rich details, for example, it captures "anti-aging properties", which is relevant to "free radicals" in the gold summary. Detailed analysis shows that 91\% of all produced tokens by our model are from the extracted segments -- critical for faithful summarization. The words generated in between allow cohesive fusion. Examples from the BreastCancer dataset can be found in the Appendix. 
Additional model outputs with human evaluation scores can be found in Appendix.

\textbf{${P(C|Z)}$ Variants}
To further understand our model, we implement an alternative content selection method, using a supervised classification module (implementation details described in Appendix 
. Table \ref{tab:automatic-ablation} reports the results. Being an extract-compose variant, the supervised model (first row) produces faithful summaries (KG(I)).  However, our Policy Network's joint selection and ability to explore span combinations with guidance from gold structure rewards and the Meteor score, lead to an improved performance on KG(G), Diversity and Meteor. Additionally, on human evaluated relevance, the Policy Network approach scores a higher 3.19 while the supervised extraction variant scores 2.5. We also observe the importance of $\mathcal{R}_m$ for KG(I) and Meteor, and $\mathcal{R}_d$ for Diversity. 

\begin{table}[!htbp]
\centering
\scalebox{0.95}{
\begin{tabular} {p{3cm}c|c|c|c|c} 
\midrule
Model    & Me     & KG(G) & KG(I) & Di    & Co    \\

\midrule

Select w/ Sup.         & 7.6   & 28 & 84 & 75 & \textbf{88} \\
Policy (w/o ${R}_m$, ${R}_d$) & 8.4 & 36 & 83 & 70 & \textbf{88}\\
Policy (w/o ${R}_m$)          & 8.2   & 33 & \textbf{89} & \textbf{91}  & \textbf{88} \\
Policy (full $\mathcal{R}$)           & \textbf{8.7}  & \textbf{40} & 88 & 90 & 83 \\
\bottomrule
\end{tabular}
}
\caption{Automatic evaluation of extract-compose model's variants on HealthLine. The best results are in \textbf{bold}.}
\label{tab:automatic-ablation}
\end{table}
\textbf{Comparison with Extractive Summarization}
To understand the benefits of fine-span extraction followed by fusion, we compare our model with an extractive summarization system (implementation details in Appendix 
. 
Our model achieves 40\% KG(G), while the extractive summarization system achieves 26\%. Our \emph{extract-compose} approach performs strongly since: (1) We extract knowledge pieces more precisely by selecting key spans instead of complex complete sentences from scientific abstracts; (2) RL model's content selection jointly through multi-steps, and (3) The Text Fusion module consolidates knowledge pieces, which may otherwise remain incomplete due to linguistic phenomena, such as coreference and discourse. 

Even unsupervised methods like TextRank ~\citep{mihalcea2004textrank} are not particularly applicable when we need to select key spans amongst multiple candidates. Instead our model is able to capture relevant, readable and coherent pieces of text by utilizing guidance from distant supervision and the use of a domain specific language model for fusion.

\section{Conclusion}
High impact datasets, content selection, faithful decoding and evaluation are open challenges in building multi-document health summarization systems. First, we propose two new datasets for \emph{Nutri-bullets}. Next, we tackle this problem by exploiting annotations on the source side and formulating an extraction and composition method. Comprehensive human evaluation demonstrates the efficacy of our method in producing faithful, informative and relevant summaries. 

\newpage
\bibliography{aaai21}
\newpage
\appendix
\section{Appendices}
\label{sec:appendix}

\renewcommand\thefigure{\thesection.\arabic{figure}}    
\setcounter{figure}{0}

\renewcommand\thetable{\thesection.\arabic{table}} 
\setcounter{table}{0}

\subsection{BreastCancer Case Study}
\label{appendix:bc case study}
In Table A.1, we show the various system outputs on BreatCancer. Like in HealthLine, Transformer, GraphWriter and copy-gen fail to generate meaningful summaries. They also tend to repeatedly generating similar text. Our extract-compose (BLM) model correctly identifies the benefit of whole grains for BreastCancer, as well as the crucial nutrition, Melatonin. We also found that in such a region of low-resource data, tranformer based models even suffer from generating grammaratically correct text, and start repeating text segments. GraphWriter even repeated certain text span 5 times in the examples. 

\begin{table*}[!h]
\centering
\footnotesize
\label{tab:casestudy-bc}
\begin{tabularx}{\textwidth}{lX}
\toprule
Food     & oats \\
\midrule
\textit{Gold }               & \textit{Diets rich in soluble fiber and whole grains have been found to be protective against breast cancer in some studies and not associated with breast cancer risk in others. Oats are a source of melatonin, which has been shown to be associated with lower risk of breast cancer. A 2018 study reported that Finnish women with high oatmeal consumption had a lower risk of breast cancer.}\\

Ours (blm)                & Two recent meta-analyses reported an inverse association between dietary fiber and whole grain intake and the risk of colorectal cancer. Evidence from previous reviews is supportive of the hypothesis that whole grains may protect against various cancers. Melatonin enhances the apoptotic effects and modulates the changes in gene expression induced by docetaxel in MCF human breast cancer cells. Results from clinical trials and multiple in vivo and in vitro studies point to melatonin as a promising adjuvant molecule with many beneficial effects when concomitantly administered with chemotherapy . Oats contains selenoproteins ; it has been suggested that several selenoproteins may be protective against cancer.  \\

Ours (trans.) & A study that examined the anti-cancer activity of eight brown rice phenols <unk> acid, p-coumaric acid, ferulic acid, sinapic acid, vanillic acid, <unk> acid, and <unk> found that ferulic acid and <unk> each were effective in reducing the colony-forming ability of triple negative breast cancer cells. Brown rice has been shown to inhibit the growth and proliferation of mouse mammary tumors. Rice bran and rice bran oil are good sources of the <unk> <unk> form of vitamin E, which has been shown to inhibit important steps in metastasis in both <unk> and triple negative <unk> breast cancer cells. An anthocyanin-rich extract from black rice reduced the viability of several breast cancer cell lines in another study. Phytic acid <unk> in rice <unk> has been shown to induce marked growth inhibition in breast cancer cells without harming normal cells. The same study found that the extract significantly suppressed tumor growth and angiogenesis in mice implanted with HER2/neu overexpressing <unk> tumors. \\

Transformer  & Consumption of sweet potato was associated with decreased risk of kidney cancer in one Japanese study and decreased risk of gallbladder cancer in an Indian study. For <unk> delphinidin has been shown to induce apoptosis <unk> cell <unk> in <unk> breast cancer cells and to enhance the treatment effects of Herceptin. Purple sweet potatoes <unk> commonly consumed in Hawaii and parts of <unk> have high levels of anthocyanins, including delphinidin and <unk> which has been shown to possess both chemo preventive and chemotherapeutic activity.   \\

Copy-gen            & While one carefully designed study that the A association between the enterolactone in the effects of the animal which in carcinogen-induced diet of carcinogen-induced diet and other of breast cancer. While the case of several studies evaluate association impact selenium or an risk of breast cancer are found that the growth. While iron of one other selenium could benign high high levels of iron of starch ingredients, found contains any growth of several studies have been shown to increase of metastasis in several breast However, improved women.   \\

Graph-writer       & has been found to be associated with lower risk of breast  copper breast conditions6 in women with breast conditions10 study found that women with high levels of  molasses  in the risk of breast  copper breast conditions6 in women with a high levels of  molasses  \textbf{$\dots$ $\dots$}  copper breast conditions6 in one study reported that women with a high intake of  molasses  in the risk of breast  copper breast conditions6 in women with breast conditions10 study found that consumption of  molasses  was associated with lower. \\
\bottomrule
\end{tabularx}
\caption{Example outputs of various systems on BreastCancer dataset}
\end{table*}

\subsection{Human Evaluation Examples}
\label{appendix:human-eval}
Table A.2 presents examples of various system outputs on HealthLine, together with the human judgement scores. One interesting obeservation is that the fluency metric does not correlate to other content based metrics. For instance, in the example for \textit{dark chocalate} and \textit{whole grain}, the most fluent outputs are least faithful to the input. 

\begin{table*}[!h]
\vspace{-10pt}
\centering
\footnotesize
\label{tab:morehl}
\begin{tabularx}{\textwidth}{p{1.8 cm}Xc}
    \toprule
      & output text & ratings \\
    \midrule
    Food & dark chocalate & \\
    \midrule
    \textit{Gold} & \textit{Dark chocolate improves several important risk factors for disease. It lowers the susceptibility of LDL to oxidative damage while increasing HDL and improving insulin sensitivity.}  & NA \\
    Ours (BLM)          & Dark chocolate and cocoa powder are flavanol-rich dark chocolate that decrease blood pressure (BP) and insulin resistance in healthy subjects. & (3, 3, 3) \\
    Ours (trans.)  & Dark chocolate may have anti-inflammatory effects for some people, but more research is needed. & (1, 4, 4) \\
    Transformer & Studies suggest that dark chocolate and dark chocolate can increase the risk of heart disease and diabetes.  & (1, 2, 2)  \\
    Copy-gen & Beef contains cocoa are high in antioxidants that may help protect against chronic health and reduce your risk of heart disease. &(1, 1, 3) \\
    GraphWriter & The type of insoluble fiber found in powder powder may help prevent weight gain and reduce the risk of chronic diseases, such as heart disease and cancer. &(1, 1, 3) \\
    \midrule
    \midrule
    Food & whole grain &  \\
    \midrule
    \textit{Gold} & \textit{Whole grains may help prevent colorectal cancer, one of the most common types of cancer. Still, research on whole grains' anticancer effects is mixed.}  & NA\\
    Ours (BLM)          & Whole grains containing fiber. The role of dietary fiber on the risk of colon and rectal cancer has been investigated in numerous studies, but findings have been inconsistent. The whole grains are observed between intake of total fiber and the risk of colon cancer per an incremental increase of 10 g day. & (4, 3, 4) \\
    Ours (trans.)  & Grain's soluble fiber may promote gut health and reduce your risk of colon cancer.  & (3, 4, 3) \\
    Transformer & Whole grains, such as blueberries, are likely safe to eat with no serious side effects.  & (1, 1, 3)  \\
    Copy-gen & The tannins and saponins in whole grains may inhibit the risk of heart disease. &(1, 1, 4) \\
    GraphWriter & The type of insoluble fiber found found that foods may help fight oxidative damage caused by colon cancer blood sugar levels. however, more research is needed to determine whether it has similar effects. &(3, 3, 2) \\
    \midrule
    \midrule
    Food & pear &  \\
    \midrule
    \textit{Gold} & \textit{Eating pears regularly may help you feel full because of their high amounts of water and fiber. In turn, this may help you lose weight.}  & NA \\
    Ours (BLM)          & Pears and fruit to reduce energy consumption and body weight over time. & (4, 4, 3) \\
    Ours (trans.)  & Studies suggest that eating pears may increase feelings of fullness and help weight loss.  & (4, 4, 4) \\
    Transformer & Pears may help lower blood pressure by lowering blood pressure and improving cholesterol levels.  & (2, 2, 2)  \\
    Copy-gen & Some studies have found that eating pears fruits can lead to blood blood sugar levels, moderate blood pressure. &(2, 2, 2) \\
    GraphWriter & The type of insoluble fiber found in oats oats may help reduce blood sugar levels, and lower blood pressure and cholesterol levels. &(1, 1, 2) \\
    \bottomrule
\end{tabularx}
\caption{Example outputs and their human ratings (faithfulness, relevance, fluency) of various systems on HealthLine.}
\end{table*}

\subsection{Data collection}
\label{appendix:data}

\paragraph{Summary Example from healthline.com}

\begin{figure}[h]
\centering
\includegraphics[width=3.2in]{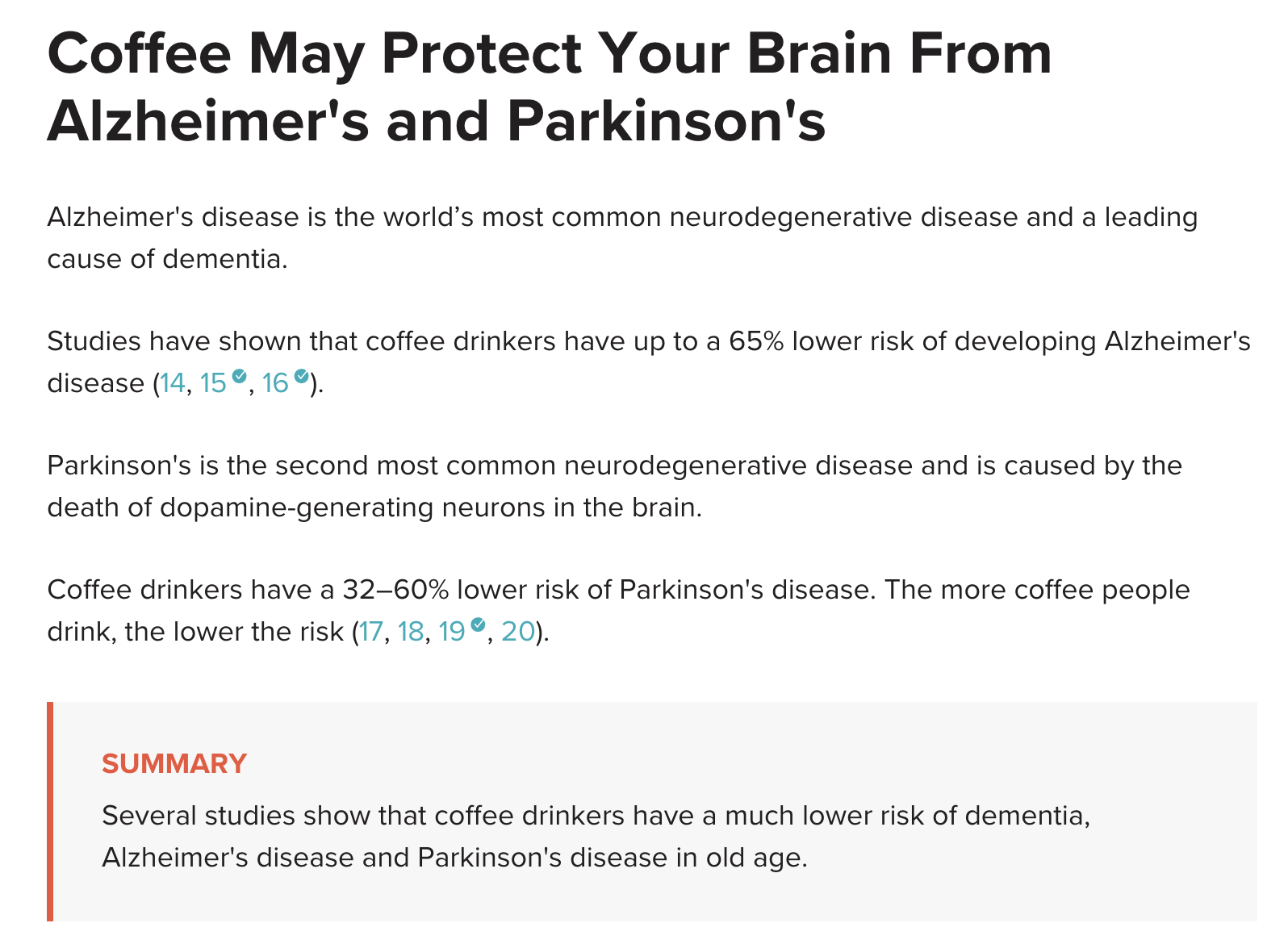}
\caption{Screenshot of a session in an HealthLine web article with title \emph{Coffee — Good or Bad?}. It shows how Pubmed articles pair with a summary.}
\label{fig:hl-coffee}
\end{figure}

Figure~\ref{fig:hl-coffee} shows a section of a typical HealthLine web article, \emph{Coffee — Good or Bad?} (https://www.healthline.com/nutrition/coffee-good-or-bad). The HealthLine article contains multiple sections, with each section talking about different aspects of coffee. As shown in the screenshot, a session contains a session title, \textit{coffee may protect your brain from Alzheimer's and Parkinson's}, a small paragraph with Pubmed articles citations and a summary box. We take the Pubmed articles and the summary text to form the parallel corpus. The unique structure of multiple section article allows us to evaluate the \textit{informativeness} and \textit{contrastiveness} of a model. 

\begin{figure}[h]
\centering
\includegraphics[width=2.8in]{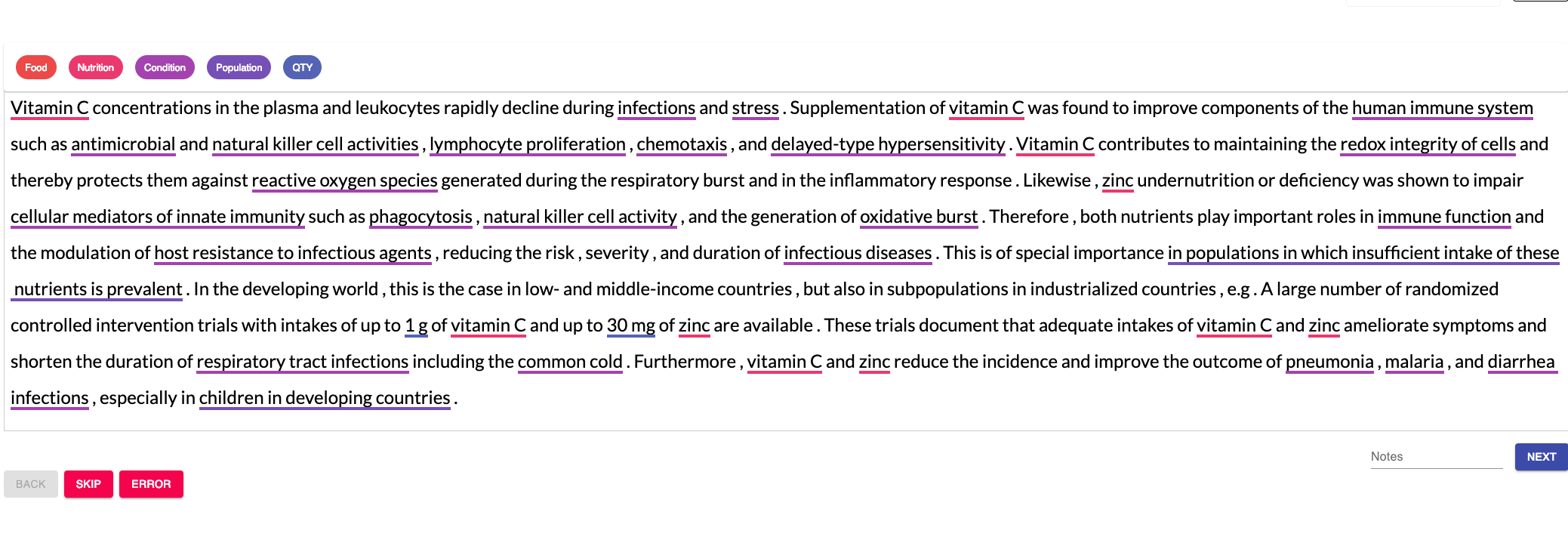}
\caption{User interface for entity highlight annotation}
\label{fig:entity-interface}
\end{figure}

\paragraph{Annotation instructions}
During the food entity annotation, we provide workers with the definition of each entity as following: 
\begin{itemize}  
\setlength\itemsep{0.005em}
    \item Food: Entity which is edible and commonly consumed by people. E.g banana, cocoa, coconut oil, etc
    \item Nutrition:  Nutrient found inside the food. E.g starch, protein, fiber, etc
    \item Condition: A bodily condition, or a body state which has a  health impact on human/animal. E,g blood sugar level, cancer. Can be something broad like ‘diseases’, ‘chronic conditions’ too.
    \item Population: Entity on which study was performed or was applicable. E.g humans, men, children, women, rats, etc
    
\end{itemize}
An example interface of entity annotation is shown as in \ref{fig:entity-interface}. 

\begin{figure}[h]
\centering
\includegraphics[width=2.9in]{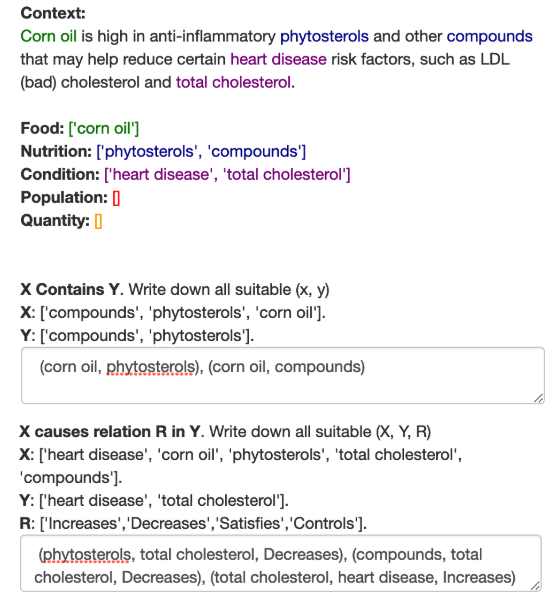}
\caption{User interface for entity highlight annotation}
\label{fig:relation-interface}
\end{figure}

For entity relationship annotation, the turkers are presented a piece of text as well as the existing entities from that text, as shown in Figure~\ref{fig:relation-interface}. we have the following instructions for the workers. 
\begin{itemize}
\setlength\itemsep{0.005em}
    \item Please write down all suitable pairs, following the format of ‘Expected Output’ in the examples below. (X1, Y1), (X2, Y2), etc or (X1, Y1, R1), (X2,Y2, R2), etc
    \item A relation type may have many or zero entity pairs. If there is no pair, please put NA.
    \item Possible entities values of X and Y is shown in the box heading, e.g (X: chocolates, cocoa, flavanols; Y: cocoa, flavanols) means X should take one of (chocolates, cocoa, flavanols) and Y should be one of (cocoa, flavanols).
    \item Please only use the entities shown in the headings. (if something is missing/wrong, please ignore it) Note that, X and Y should be different entities.
    \item In causes-relationship annotation, in case of an ambiguity between food/nutrition which can cause a condition. Always pick the most fine entity. i.e the nutrition to have it cause the condition. (see example 1)
    \item In causes-relationship annotation, in case food/nutrition directly impacts condition1 which further impacts condition2. Please annotate it as: (food/nutrition, condition1, cause), (condition1, condition2, cause). In such a specific case, we won’t annotate (food/nutrition, condition2, cause)
\end{itemize}

\subsection{Implementation Details}
\label{section:appendix_implementation}
\paragraph{Our Model implementations}
For extraction model, we use a 200 dim bidirectional LSTM network \citep{hochreiter1997long} and 300-dim FastText \citep{bojanowski2017enriching} embeddings trained on Pubmed articles. For policy network implementation, each text span is encoded using 50 dimensional FastText embeddings. We train our the entire data for up to 50 epochs, using Adam optimizer \citep{kingma2014adam} and a learning rate of 0.05.
\newline
 For BLM, we adapt the implementation of \citet{shen2020blank}, and the vocabulary of each categorical \textit{blank} is shown in Table~\ref{table:blank_language_model_vocabulary}. 
While the last 2 categorical blanks have a limited vocabulary, they can still give birth to new "vanilla blanks" to fill in more words from the entire vocabulary. In practice we find this to be quite effective for our task specific infilling.

\begin{table}
\footnotesize
\begin{tabular}{c|c}
\toprule
Entity Pairs       & Vocabulary Included \\ 
\midrule
Food/Nutrition\_Condition    & Entire Vocabulary \\ 
Food\_Food    & like, and, \\ 
Food\_Nutrition & contain(s),include(s),consist(s) \\
\bottomrule
\end{tabular}
\caption{Details of vocabulary constraints for different entity pairs}
\label{table:blank_language_model_vocabulary}
\end{table}

\paragraph{Supervised Selection Model} The gold structure tuples are used as supervision to classify amongst rationale candidates, the most relevant and important. The model is a logistic regression classifier, which selects 2 per test instance -- same as the average number of rationales selected by our policy network on HealthLine.

\paragraph{Extractive Summarization}
We use the strong BERT based extractive summarization model, off the shelf \url{https://github.com/dmmiller612/bert-extractive-summarizer} as created in ~\citet{miller2019leveraging}. As we tend to pick an average of two rationale spans in the HealthLine dataset, we make this model select two sentences as well.

\end{document}